\documentclass[final]{cvpr}

\usepackage{times}
\usepackage{epsfig}
\usepackage{graphicx}
\usepackage{amsmath}
\usepackage{amssymb}
\usepackage{booktabs}
\usepackage[para,online,flushleft]{threeparttable}
\usepackage{support-caption}
\usepackage{subcaption}
\usepackage{tabularx,rotating}
\usepackage[utf8]{inputenc} % allow utf-8 input
\usepackage[T1]{fontenc}    % use 8-bit T1 fonts
\usepackage{paralist}

\usepackage{wrapfig}
\usepackage{url}            % simple URL typesetting
\usepackage{booktabs}       % professional-quality tables
\usepackage{amsfonts}       % blackboard math symbols
\usepackage{amsmath}
\usepackage{nicefrac}       % compact symbols for 1/2, etc.
\usepackage{microtype}      % microtypography

\usepackage{multirow}
\usepackage{array}
\usepackage{amsfonts}
\usepackage[utf8]{inputenc}
\usepackage[para,online,flushleft]{threeparttable}
\usepackage{multirow}

\usepackage{silence}
\newcommand{\forloop}[5][1]{
    \setcounter{#2}{#3}%
    \ifthenelse{#4}{#5\addtocounter{#2}{#1}%
    \forloop[#1]{#2}{\value{#2}}{#4}{#5}}%
    {}}

\newcounter{crcounter}

\newcommand{\mcrot}[4]{\multicolumn{#1}{#2}{\rlap{\rotatebox{#3}{#4}~}}} 
\newcommand*{\twoelementtable}[3][l]%
{%  
    \begin{tabular}[t]{@{}#1@{}}%
        #2\tabularnewline
        #3%
    \end{tabular}%
}
\usepackage[pagebackref=true,breaklinks=true,colorlinks,bookmarks=false]{hyperref}

\def\cvprPaperID{7303} % *** Enter the CVPR Paper ID here
\def\confYear{CVPR 2021}
\begin{document}

%%%%%%%%% TITLE
\title{LORCK: Learnable Object-Resembling Convolution Kernels}
\author{Elizaveta Lazareva*, Oleg Rogov*, Olga Shegai**, Denis Larionov**, and Dmitry V. Dylov*\\
*Skolkovo Institute of Science and Technology\\
**FSBI National Medical Research Radiological Center \\of the Ministry of Health of the Russian Federation \\
Moscow, Russia\\
\texttt{\{elizaveta.lazareva, o.rogov, d.dylov\}@skoltech.ru}\\
% \texttt{olga.shegai2020@outlook.com}
}

\maketitle

%%%%%%%%% ABSTRACT
\begin{abstract}
   Segmentation of certain hollow organs, such as the bladder, is especially hard to automate due to their complex geometry, vague intensity gradients in the soft tissues, and a tedious manual process of the data annotation routine. Yet, accurate localization of the walls and the cancer regions in the radiologic images of such organs is an essential step in oncology. To address this issue, we propose a new class of \emph{hollow kernels} that learn to 'mimic' the contours of the segmented organ, effectively replicating its shape and structural complexity. We train a series of the U-Net-like neural networks using the proposed kernels and demonstrate the superiority of the idea in various \emph{spatio-temporal} convolution scenarios. Specifically, the dilated hollow-kernel architecture outperforms state-of-the-art \emph{spatial} segmentation models, whereas the addition of \emph{temporal} blocks with, e.g., Bi-LSTM, establishes a new multi-class baseline for the bladder segmentation challenge. Our spatio-temporal model based on the hollow kernels reaches the mean dice scores of 0.936, 0.736, and 0.712 for the bladder's inner wall, the outer wall, and the tumor regions, respectively. The results pave the way towards other domain-specific deep learning applications where the shape of the segmented object could be used to form a proper convolution kernel for boosting the segmentation outcome.  
\end{abstract}

%%%%%%%%% BODY TEXT
% \begin{figure}
%     \centering
%     \includegraphics[width=0.5\textwidth]{Figures/temp_promo.PNG}
%     \caption{}
%     \label{fig:owl_promo}
% \end{figure}
\section{Introduction}

Timely and minimally invasive diagnosis of bladder cancer (BC) is one of the most pressing problems of modern oncology. For this cohort of patients, the treatment success is closely related to the speed and the adequacy of the tumor stage determination~\cite{Milowsky2016}. One of the most powerful visualization tools for the BC patients is Magnetic Resonance Imaging (MRI), which has immediately become the go-to solution for precise localization of the tumors following the first publications on the MRI use in the urinary tract diagnostics circa 1986~\cite{Lee1986_mri}. The consequent improvement of the MRI modality has enabled many diagnostic breakthroughs both in cancer research and in clinical practice~\cite{glance2018,review2019}. 

Yet, the MRI is still considered an imperfect modality for BC imaging, with multiple types of artifacts impeding the adequate assessment of the bladder’s structures~\cite{Sin2017}.
A non-uniform filling and stretching, intestine peristalsis, body movements, and chemical shifts during the MRI recording are among the major factors that cause the MRI artifacts to appear in the images of the bladder~\cite{Sin2017}.
Besides, the thickness of the healthy wall of the bladder could range from 2 mm with a moderate filling to 8 mm in a collapsed condition~\cite{Ahn2015}, complicating the automation of the image analysis and the segmentation~\cite{Jung2004MRI}.

These impediments and the lack of publically available well-annotated datasets make the bladder one of the most dismissed organs within the medical image computing community.
While some state-of-the-art (SOTA) deep learning-based models \cite{Ronneberger2015, enet} were reported to successfully localize the outer and the inner walls of the bladder, obtaining proper segmentation of the rarely present neoplasms (cancerous regions) remains a challenging problem~\cite{Dolz2018}. In this work, we aspired to develop a universal segmentation model capable of localizing all tissues involved in the MRI scans of the BC patients using a single architecture. Motivated by the convolutions from the imaging physics, where the shape of a camera's aperture can control the very mechanism of image formation, we wanted to see if some \emph{shape-aware} kernels could similarly control the quality of 'clustering' of the groups of pixels to aid the multi-class segmentation. As a result, these shape-aware kernels, which we coined 'the hollow', yield a new baseline for the multi-class segmentation in the bladder.

\paragraph{Contributions} 
\begin{enumerate}
\item We propose \textit{learnable shape-aware kernels} for convolutional neural networks (CNN).
\item We show that, if the shape of an object is hollow (e.g., the bladder), the learnt kernels are also \emph{hollow}, which boosts the segmentation quality of the CNN-based bladder models.
\item We examine segmentation outcome of a dozen of \emph{spatial} (e.g., dilated \cite{Dolz2018}) and \emph{temporal} (e.g., Bi-LSTM\cite{bai2018recurrent}) CNNs equipped with our kernels.
\item Built in a \emph{spatio-temporal} configuration, a basic U-Net-like architecture, boosted with the hollow kernel convolutions, becomes a new SOTA in the multi-class bladder segmentation challenge.
\end{enumerate}

\section{Related work}
The general problem of image segmentation for biomedical tasks is well explored~\cite{Zhou2019}. 
Significant advances were made with the following architectures based on the Convolutional Neural Networks (CNN): U-Net \cite{Ronneberger2015}, E-Net \cite{enet}, Stack U-Net \cite{Sevastopolsky2019}, U-Net++ \cite{zhou2018unetPlusPlus}, MD U-Net \cite{zhang2018mdunet}, 3D U-Net \cite{iek2016_3dUnet}, V-Net \cite{milletari2016_Vnet}, etc. Nonetheless, the subject continuously attracts more and more attention because of the multiple challenges that certain nosologies still face today, ranging from the poor data availability~\cite{prokopenko2019unpaired}, to the lack of robust low-contrast segmentation methods, and to the complexity and the long duration of the manual annotations required for certain organs and imaging modalities~\cite{chowdhury2017blood,zhou2018unetPlusPlus}.

One such direction --- somewhat dismissed by the expert annotators --- is the hollow organ segmentation in general, and the urine bladder in particular. For the problem of segmentation of the bladder structures, several \emph{spatial} modifications based on the use of sparse convolutional kernels for U-Net models were recently  proposed in~\cite{Dolz2018}. The authors proposed two architectures in which the convolution kernels were dilated at different extent in a specific order, showing the SOTA results on their dataset.

Popular approach for the image sequence data (e.g., a video or an MRI stack) is to combine a CNN architecture with a recursive block such as Long-Short-Term Memory (LSTM) \cite{hochreiter1997long} for capturing the inter-slice context. Generalizations using convolutional LSTMs (ConvLSTM) \cite{xingjian2015convolutional}, Gated Recurrent Units (GRUs) \cite{cho2014properties}, and special bottlenecks \cite{poudel2016recurrent} -- have been exploiting the similar concept. 
Among such `\emph{slice-remembering}' architectures, one of the most powerful ones is the method that extracts feature maps from the adjacent MRI images and then uses Bi-Directional Convolutional LSTM (BiLSTM \cite{bai2018recurrent}) to correct them via predicting the input images \cite{chen2016combining}. In the video segmentation domain, these recurrent networks have recently been superseded by \emph{temporal} models \cite{jain2018fast}, \cite{li2018low}, which significantly accelerated the prediction without sacrificing the segmentation accuracy. Extension of this \emph{temporal} methodology to the biomedical domain has not been reported yet, let alone a combinations of them with the \emph{spatial} dilation methods described above.

Another line of development relates to \emph{learnable kernel} methods which proved instrumental in areas ranging from audio processing \cite{AudioUnet2019} to medical imaging and microscopy\cite{pronina2019microscopy}. Several recent papers report atrous convolutions \cite{chen2017rethinking} and their blocks \cite{chen2017deeplab} which eliminate the need for max-pooling and striding; as well as, progressive dilated convolutional blocks \cite{Dolz2018}, which boost the receptive field and scale the number of training parameters via the alternating dilation rate. However, with the far-fetched exception of Deformable Convolutional Networks \cite{dai2017deformable} (which learn patterns via offsets from the targets with adaptable kernels), no effort has been devoted to studying the segmentation performance when \emph{the contours} of the target object coincide with those of the kernel.

\section{Methods}
\begin{figure*}[ht]
\centering
    \includegraphics[width=0.9\textwidth]{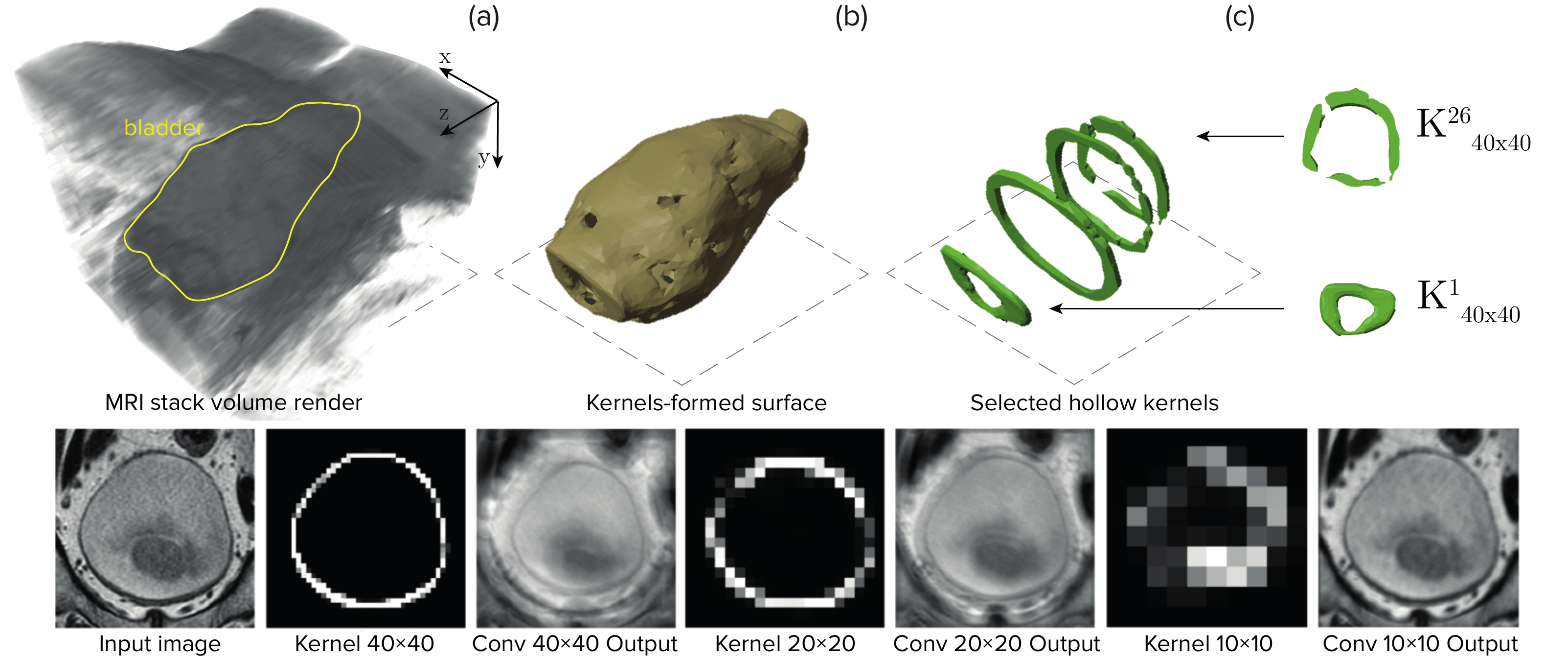} \caption{\textbf{\textit{Top}: The bladder in 3D and the hollow kernels which mimic the shape of the organ.} Shown are the original MRI stack volume (with added transparency), segmented outer wall of the bladder, and several of the hollow kernels with the dimensions of 40$\times$40 used to build our \emph{spatio-temporal} segmentation model (described in Section~\ref{A2}). \textbf{\textit{Bottom}: Intuition behind the hollow kernels:} convolution with them emphasizes the contours of the organ along with the tissue borders in the vicinity. The effect depends on the scale of the convolution kernel.}
\label{fig:promo}
\end{figure*}
In this section, we introduce a single architecture with \emph{learnable spatio-temporal hollow kernels} capable of performing multi-class segmentation (the inner bladder wall, the outer wall, and the tumors). Our first step is a modification of the U-Net realization from  \cite{Dolz2018} to introduce the \emph{spatial} (or the lateral) dependence via our \emph{shape-aware} kernels, with the original dilutions kept intact and their rates doubled in each deeper layer. Next, we propose further enhancement of the spatial hollow kernel convolutions by adding the \emph{temporal} inter-relation between the slices.
\paragraph{Spatial Encoding: Hollow Kernels} Many anatomical organs, and the bladder in particular, encompass hollow three-dimensional structures within the soft tissue walls (see Fig.~\ref{fig:promo}) which frequently appear vague in the MRI data, lacking the necessary contrast. These very bagel-like shapes of the organ have motivated us to pre-train kernels which themselves would resemble these spatial features (`bagels'). As could be seen in Fig.~\ref{fig:promo} (bottom row), the boundaries of the object are highlighted and even framed when the scale of the convolution kernel is \emph{optimal}. 
 
Formally, we will denote the hollow kernels as $\mathbf{K} \in \mathbb{R}^{K \times K \times C}$ for an image with $C$ channels. The observation that the ratio of the linear kernel size $K$ to its width $W_{wall}$ for a given hollow object plays a key role in the boundary highlighting effect is actually physically intuitive, being a type of stochastic denoising problem~\cite{dylov2011nonlinear}.
Specifically, if $K\ll W_{wall}$, the usual blurring of the image occurs. Whereas, at a certain point, the increase of the kernel size produces a thin hollow structure in the output. At the optima $K \sim W_{wall}$, the boundary of the convolved object coincides with the original one. Further expansion of the kernel scale reverses the constructive effect of the convolution and, instead, deteriorates the visibility of the borders (see the ``\texttt{Conv 40$\times$40 output}'' \textit{vs.} ``\texttt{Conv 20$\times$20 output}''). Naturally, if introduced within a CNN, such an effect could be expected to boost the segmentation performance of those objects in the image, the shape of which is \emph{homothetic} to the learnt kernel.
% Our idea is to introduce such convolutions into the CNN model. However, a hollow core stack leads to a loss of spatial information about central pixels. To solve this problem, put a point in the center of mass of the nucleus

\begin{figure*}[t]
    \centering
    \includegraphics[width=1\textwidth]{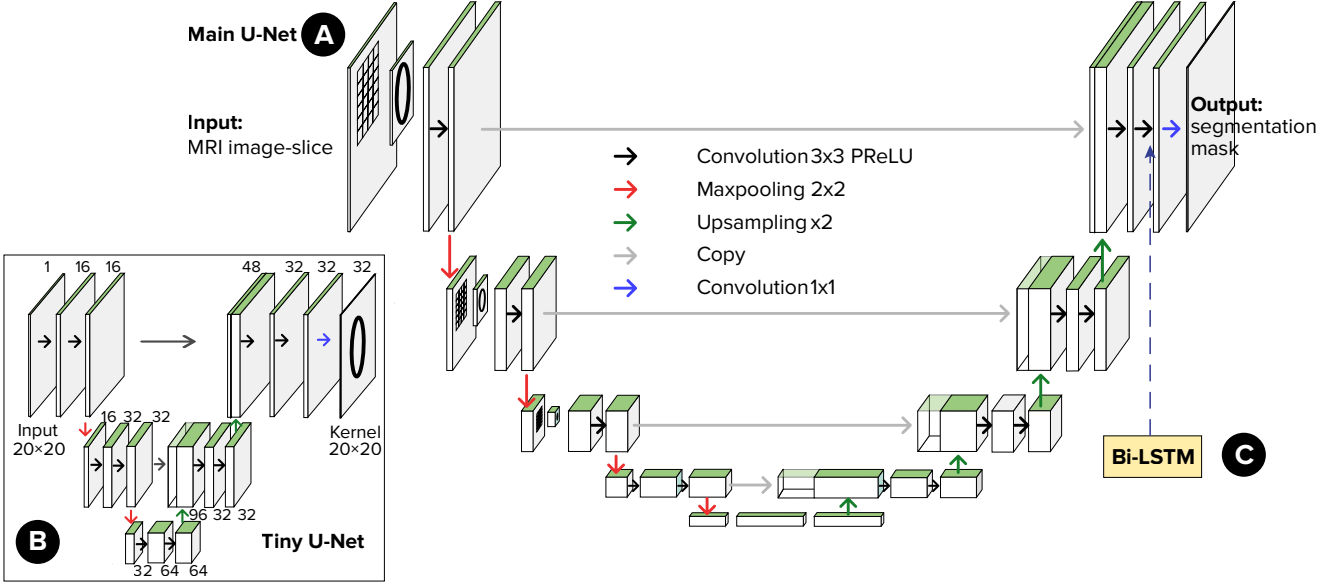} \caption{\textbf{Proposed \emph{spatio-temporal} architecture with \emph{hollow} convolutions:} (A) Main U-Net module; (B) Tiny U-Net used for kernel generation; (C)  Bi-directional Convolutional LSTM block.} 
    \label{fig:unet_hollow}
\end{figure*}

\paragraph{U-Net with the Hollow Kernels}
To fulfill the vision, we construct a U-Net architecture, with the first convolution layers of the first three blocks leveraging the hollow kernels (see Fig.~\ref{fig:unet_hollow} (A)). 
Although the exact optimization task of matching the sizes of the kernel and the object remains to be solved theoretically, herein, we limit our consideration to an optimization of the network hyperparameters that describe these kernels. Namely, we consider brute force optimization over their spatial sizes in each layer, the dilation rate, as well as the intensity fluctuations within the given 'bagels' of the kernels.

In our experiments, we observed that the gradual increase of the kernel size similarly to the configuration of the U-Net Dilated is indeed efficient. Namely, the first block kernel can be chosen as 10$\times$10 ($\mathbf{K}_{10\times 10}$), the kernel of the next one is then 20$\times$20 ($\mathbf{K}_{20\times 20}$), and the last block entails kernels of the size 40$\times$40 ($\mathbf{K}_{40\times 40}$). To preserve the global features present at the deeper layers, we purposely upsampled the definition matrices to initialize these kernels (i.e., made them larger with some margin). The exact recipe for `learning' these kernels could follow the following two intuitive approaches.
\paragraph{Kernel Initialization Approach 1: Kernel Generation by Tiny CNNs} \label{A1}
The first approach involves a model where each image’s convolution uses kernels emulating the bladder walls from the input image. Therefore, we needed to step away from the classical deep learning training paradigm where all neural network weights are evenly initialized and updated with the gradient descent to minimize the loss function. Our approach naturally assumes that it is possible to generate the convolution weights by some of the U-Net models for \emph{each} input image and, consequently, this process cannot be optimized in the classical manner.

For our goals, we propose to use multiple small CNN models which will predict the kernels needed for the main model layers. As the input, these tiny models use an image with a size equal to the respective kernel size that one wants to obtain. Thus, we used shallow U-Net models (Fig. \ref{fig:unet_hollow} (B)), hereafter ``Tiny U-Nets''.
To elaborate, let us consider some convolution layers with the hollow kernels in the proposed architecture of the Fig.~\ref{fig:unet_hollow}. We define such convolution as a differentiable function of two input variables $y = f(x, w)$, where $w$ is a kernel, obtained from the Tiny U-Net model, and $x$ is an output from the previous layer of the main U-Net model. Let $L(y)$ denote the loss function which represents all layers which sequentially follow the current one.
To do the backward step, we compute the gradients of $L$ with respect to $x$ and $w$:
\begin{equation}\label{d1}
dx =  \nabla_x L (f(x, w)),\quad  
% dx =  \nabla_x L = \nabla_x L (f(x, w))  
% \end{equation}
% \begin{equation}
dw =  \nabla_w L (f(x, w)) 
% dw =  \nabla_w L = \nabla_w L (f(x, w)) 
\end{equation}
Subsequently, we use~(\ref{d1}) to compute the gradients for the layers preceding the ones from the main U-Net and from the Tiny U-Net, respectively. Ultimately, three such Tiny U-Nets were trained for the architecture presented in Fig.~\ref{fig:unet_hollow} by adding the corresponding terms to the loss function. Their purpose is to initialize the weights of the three corresponding convolution layers of the main model.
\paragraph{Kernel Initialization Approach 2: Shape-Preserving Optimization}\label{A2}
The second approach aspires to create the same hollow shapes of the kernels, while relying not on three Tiny U-Nets but on some synthetic manipulations of the shape of the kernels using \emph{only one} of the Tiny U-Net outputs\footnote{Moreover, similarly to many other deep learning problems, instead of generating the small `bagel' from a separate Tiny U-Net, one could just look at several examples of the annotations merely to initialize the pixel values in the contours of the object. In our experiments, we find that these two starting points result in two statistically indistinguishable outcomes.}. Having been initialized, the digitally scaled kernels are then used in the corresponding convolution layers at different depths, assuring the proper scaling of dimensions. Every kernel is normalized after truncation which zeros out the values everywhere except for the elements forming the hollow structure. During the training, only the kernel pixels that belong to the contours of the hollow kernels are allowed to update their weights, while the other pixels remain zero. 
Specifically, the optimization of the `shape-preserving' kernels occurs according to the following update rule:
\begin{equation}\label{eq:step}
    %  w_{\eta} = w_{\eta-1} - \lambda * w_{weight \ hollow \ mask} * \nabla_W_{\eta-1} \operatorname{L}
     w_{\eta} = w_{\eta-1} - \lambda * w_{\mathcal{M}} * \nabla_{ w_{\eta-1} } L,
\end{equation}
where $w$ is the weight of the convolution layer, $L$, as before, denotes the loss function that represents all layers sequentially following the current one, $\eta$ is the step of the gradient descent, $\lambda$ is the training learning rate, and $w_\mathcal{M}$ is a binary mask of the pixels that form the hollow structure in the kernels.
We show a set of examples of such kernels in the Supplementary material (both initialized and trained).
% FROM SUPP BEGIN
\paragraph{Temporal models}\label{s:temporal}
% As noted in the main text, we also applied 
% The so-called temporal convolutions allow the inter-slice information flow (see Fig.~\ref{fig:temporal_config}).

\begin{figure*}
    \centering
    \includegraphics[width=0.75\textwidth]{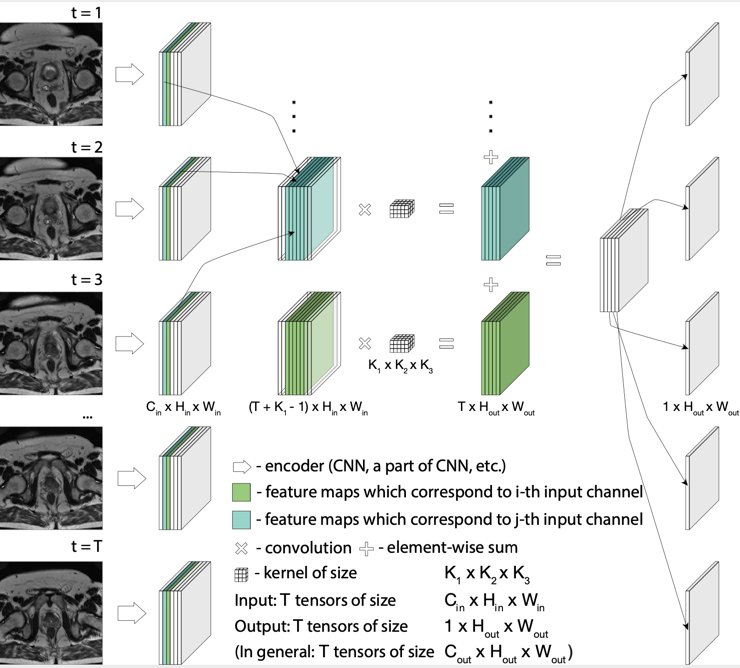}
    \caption{\textbf{Scheme of a temporal convolution with a single convolutional kernel.} On the left side, sliced boxes represent tensors of $T$ encoded input images of size $(C_{\text{in}}, H_{\text{in}}, W_{\text{in}} )$, where each slice corresponds to the certain tensor channel. Slices corresponding to the same channel number are concatenated and convolved with the given kernel. The results are summed up into a tensor of size  $(T', H_{out}, W_{out})$. In this example, transparent boxes represent paddings, because of which the resulting tensor has the same number of value of the first dimension ($T' = T$). } 
    \label{fig:temporal_config}
\end{figure*}
Let us consider the set of original sequential MRI image slices. Let $T$ denote the number of slices. Suppose that some part of U-Net model was treated with these set and the corresponding set of $T$ feature maps of sizes $(C_{\text{in}}, H_{\text{in}}, W_{\text{in}})$ were obtained. Then, given a tensor $\mathbf{x}$ of size $(N, T, C_{\text{in}}, H_{\text{in}}, W_{\text{in}})$, where $N$ is a batch size, the proposed temporal convolution can be expressed as follows:
\begin{equation}
\mathbf{x}_{N_{i}}^{C_{\text {out} j}}=\text{bias}(C_{\text{out} j})+\sum_{k=0}^{C_{\text{in}}-1} \mathcal{W}\left(C_{\text {out} j} , k\right) * \mathbf{x}_{N_{i}}^{k}
\end{equation}
where $N_{i} \in [0, N-1]$, the size of the output tensor is $(N, T^{'}, C_{\text{out}}, H_{\text{out}}, W_{\text{out}})$, $C_{\text {out } j} \in [0, C_{\text {out }}-1]$. $\mathcal{W}$ denotes a tensor of convolutional kernel weights of size $(C_{\text{out}}, K_1, K_2, K_3)$, where $C_{\text{out}},$ is a number of kernels and $(K_1, K_2, K_3)$ is a kernel size corresponding to the dimensions of input tensor $(T, H_{\text{in}}, W_{\text{in}})$. The value $K_1$ reflects the number of slices in a MRI sequence between which the information flows. The scheme in Fig.~\ref{fig:temporal_config} shows how the convolution with single kernel ($C_{\text{out}}$ = 1) works. 

In general, the size of output feature maps differs from the input size. The use of appropriate padding values allows to preserve the size $(H_{\text{in}}, W_{\text{in}})$ of feature maps and the number $T$ of them. 
 
In our study, we varied the number of layers in the temporal block: we examined both the block of one Temporal layer and the sequence of the Temporal layer, BatchNorm, PReLU and Temporal layer. In our experiments, we do not change the size of the feature maps tensor through the temporal block, the kernel size $K_1$ varies from 5 to 9 with corresponding paddings, $K_2$ equals to $K_3$ and varies from 1 to 5.

\paragraph{Spatio-Temporal Model: Hollow Kernel Convolutions with Bi-LSTM} 
% We describe further boosting of our CNN with the \emph{temporal} convolution techniques in the Supplementary material in detail. Herein, we will only report the best-performing out of the proposed models. Namely,
We find that the inter-slice information could be leveraged in the optimal manner via a combination of our hollow kernel models with the Bi-Directional Convolutional LSTM block~\cite{chen2016combining}, inserted immediately before the last U-Net layer (Fig.~\ref{fig:unet_hollow} (C), similarly to \cite{bai2018recurrent}). This way, a set of individual feature maps generated in the first layers of the model for each neighboring slice efficiently exchange the `adjacent' data to enhance contextual learning.  The updated feature maps are then passed through the last convolutional layer to obtain the final \emph{spatio-temporal} predictions.

\section{Experiments}
\paragraph{Data}
% \begin{table}[h]\label{tab:c_data}
%     \caption{Clinical characteristics of bladder cancer patients employed in the study.}
%     \centering
%     \begin{tabular}{cc}
%         \toprule
%         Characteristic & Value \\
%         \midrule
%         Patients, \# (\%) & 32 (100\%) \\
%         Female, \# (\%)  & 3 (9.68\%) \\
%         Male, \# (\%)  & 29 (90.32\%) \\
%         Age, Median (Range), yrs & 62 [38, 88]\\
%         \bottomrule
%     \end{tabular}
% \end{table}
\begin{table}
    \caption{\textbf{Quantitative characteristics of the dataset used in our study.}}
    \centering
    \begin{tabular}{cc}
        \toprule
        Characteristic & Value \\
        \midrule
        Series, \# & 109 \\
        Images, \# (\%) & 3783 (100\%) \\
        Images with bladder, \# (\%) & 1914 (50.59\%)\\
        Images without bladder, \# (\%) & 1869 (49.41\%)\\
        \bottomrule
    \end{tabular}
\label{qdata}
\end{table}

All experiments were conducted on the dataset collected in P. Hertsen Moscow Oncology Research Institute, containing an abdominal MRI data from 32 unique patients in DICOM format. The patient data employed in the study are presented in Table~\ref{qdata} featuring quantitative statistics. The data for each patient include several image stacks, each recorded in a specific view: sagittal, frontal, and axial. 
The manual annotation was performed for each patient, with all three classes having been labeled by two expert radiologists in Osirix Lite software (bladder outer and inner walls, and tumor regions). The data were divided into training (23 patients) and testing (9 patients) sets. Because the bladder segmentation is a very dismissed problem in the medical vision community (no publicly available datasets to merge with), we resorted to using all available data for the training/testing split and omitted separate experiments on a validation set\footnote{We openly call other research groups to merge our datasets for the common benefit (e.g., authors of \cite{Dolz2018}).}.
As a preprocessing step, we cropped the DICOM slices to 256$\times$256 pixels to localize the bladder. Images that did not contain the bladder at a given view plane were still cropped to the same size. We also applied normalization to the images using the mean (0.273) and the standard deviation (0.067), calculated on the entire dataset.

\paragraph{Training and Evaluation} 
All models were trained with the Adam optimizer for 60,000 iterations and the batch size of 8. The initial learning rate of 0.001 was multiplied by 0.1 after each 15,000 iterations.
Standard DICE metric was used to evaluate the segmentation quality.
% To evaluate the segmentation accuracy, several standard quality metrics, such as the Intersection over Union ($\operatorname{IoU}$) and the DICE coefficient ($\operatorname{DICE}$), were used.
The experiments were conducted in Python using the PyTorch framework and were run on Nvidia GeForce GTX 1080 Ti GPU with 11GB VRAM\footnote{Our code is available at: \url{https://github.com/cviaai/LORCK}}.

All models were trained with the combination of Weighted Binary Cross Entropy loss (WBCE) and DICE loss:
\begin{equation}
\label{equation_main_loss}
L[y, \hat{y}] = \alpha \ \text{WBCE}\big(y, \hat{y}\big) + (1-\alpha) \ \text{DICE}\big(y, \hat{y}\big)
\end{equation}
% WBCE
\begin{equation}
% \text{WBCE}[y, \hat{y}]
\text{WBCE} = - \sum_{c=1}^C w_c [p_c y_c \log{\hat{y}_c} - (1- y_c) \log{(1 - \hat{y}_c)}]
\end{equation}
% DICE
\begin{equation}
% \text{DICE}[y, \hat{y}]
\text{DICE} = 1 - 2 \frac{ |y \cap \hat{y}|}{|y| + |\hat{y}|}    
\end{equation}
% \begin{align}
%     \label{equation_main_loss}
%     \operatorname{loss} [y, \hat{y}] = \alpha \ \text{WBCE}\big(y, \hat{y}\big) + (1-\alpha) \ \text{DICE}\big(y, \hat{y}\big), \\
%     \notag
%     \text{WBCE} [y, \hat{y}] = - \sum_{c=1}^C w_c [p_c y_c \log{\hat{y}_c} - (1- y_c) \log{(1 - \hat{y}_c)}],   \\ \notag
% \text{DICE} [y, \hat{y}] = 1 - 2 \frac{ |y \cap \hat{y}|}{|y| + |\hat{y}|}
%     \notag
% , 
% \end{align}
where, $y$ is the target mask and $\hat{y}$ is a probability mask predicted by a model, $C$ is the number of classes, coefficients $w_c$ are the weights of classes, $p_c$ are the the positive weights of classes. 
The class weights were taken to be proportional to the number of pixels belonging to each class: $w_c$=\{4.1, 1.4, 8.7\} for the outer wall, the inner wall, and the tumor regions, respectively. The positive weight for each class was defined as an average over the ratio of the image pixels belonging to that class to the total number of all other pixels. The positive weights were $p_c$=\{25.7, 8.2, 55.0\}, correspondingly. Coefficient $\alpha$ was fixed at an empirically found value of 0.1.

%%%%%%%%%%%%%%
%%%%%%%%%%%%%%
%%%% Do NOT edit anything above this line please
%%%%%%%%%%%%%% 
%%%%%%%%%%%%%%

\paragraph{Implementation details}
Experiments with the hollow kernel models can be divided into two groups, according to the kernel initialization approach. We will now describe implementation details for both approaches.

\emph{In the first approach (A1)}, we used a U-Net model where the first convolution layer employs kernels of the size 20$\times$20, generated by the Tiny U-Net model, as shown in the Fig. \ref{fig:unet_hollow} (B). We emphasize that the main U-Net model and the kernel-generating Tiny U-Net model were always trained together. 
Let us denote the Tiny U-Net model as $u(\cdot)$, which takes the image resized to the size 20$\times$20 as an input, and the U-Net model as $U(\cdot, \cdot)$, which takes the original image and the generated kernel as input. If $x, y$ are an image and the ground truth mask, using formulas (\ref{equation_main_loss}), the loss for training is:
\begin{equation}\label{eq:A1loss1}
% \label{eq:A1loss}
    L_{1} = L\big[y, U(x, w_{20\times20}) \big] 
\end{equation}
\begin{equation}\label{eq:A1loss2}
    L_{2} = L_{1} +
    L\big[y^{OW}_{20\times20}, w_{20\times20} \big] 
\end{equation}
\begin{equation}\label{eq:A1loss3}
    L_{3} = L_{1} + L\big[y^{OW}_{20\times20}, W_{20\times20} \big] 
\end{equation}
% \begin{align}  
%     \operatorname{L1} = \operatorname{L}\big(y, F(x, w_{20\times20}) )\big), \\
%     % \notag
%     \operatorname{L2} = \operatorname{L}\big(y, F(x, w_{20\times20}) )\big) +
%     \operatorname{L}\big(y^{OW}_{20\times20}, w_{20\times20}) )\big), \\
%     \operatorname{L3} =  \operatorname{L}\big(y, F(x, w_{20\times20}) )\big) + \operatorname{L}\big(y^{OW}_{20\times20}, W_{20\times20}) )\big), \\
% \end{align}
% \begin{align}  
%     L = L_{1}\big[y, U(x, w_{20\times20}) )\big] +
%     L_{2}\big[y^{\text{OW}}_{20\times20}, w_{20\times20} ) )\big]
%     \label{eq:A1loss}
% \end{align}
where $w_{20\times20} = u(x_{20\times20})$ is the predicted kernel, with 20$\times$20 meaning that the object was resized to the size 20$\times$20, and $y^{\text{OW}}$ is the ground truth binary mask for the bladder outer wall (OW) class. Equations~(\ref{eq:A1loss1},\ref{eq:A1loss2},\ref{eq:A1loss3}) were further experimented with to consider several natural combinations of loss functions: main U-Net output and the all-class masks ($L_{1}$); $L_{1}$ plus Tiny U-Net output for OW and the $y^{\text{OW}}$ mask ($L_{2}$); $L_{1}$ plus class-weighted ($W_{20\times20}$) output of Tiny U-Net and the $y^{\text{OW}}$ mask ($L_{3}$). Set up this way, the Tiny U-Net network learns to predict kernels with the hollow objects that resemble the shape of the bladder's outer wall.
% $W_{20\times20} = \sum_i a_{i}\times w_{20\times20}^{i}$ and $a_i$ are the individual weights.

\emph{In the second approach (A2)}, one applies optimization that preserves the shape of the convolution kernels. To implement it, we start with the weight initialization, where we take a random ground truth mask of the bladder outer wall and fill the mask with the values sampled from the normal distribution.
Motivated by the efficiency of the Dilated U-Net, we took it as the model foundation (and as our main benchmark). We then gradually moved towards training the final model with the three convolution blocks employing the hollow kernels. Initially, only the first convolution block of the U-Net Dilated architecture was modified; then, only the second, etc. Finally, we trained the entire model with the three modified blocks. 
% \ref{tab:three_blocks}
The Notation Table in the Supplement shows the parameters of the first three blocks' convolution layers for the first 3 depth levels of the Dilated U-Net model in the extensive line of our experiments. 
In brief, the following notation is used here: A2 refers to approach 2; the first digit in the configuration number corresponds to the number of blocks in the U-Net Dilated model with the hollow kernel convolutions (e.g. A1 Config~1.1 means that only the first block or the first two convolution layers were modified); the last digit in the configuration number is entered to distinguish specific details of the three architectures related to A2. 
The first one uses $10\times10$ hollow kernels and an increasing dilation rate (similarly to Dilated U-Net). The second architecture employs kernels of different sizes: $10\times10$ for the first block, $20\times20$ for the second block, and $40\times40$ for the third one.  Finally, the same kernel with constant dilations was used in the third architecture.

\emph{In the spatio-temporal model,} the best-performing U-Net with the hollow kernels was combined with the Bi-LSTM block. We varied the block's parameters, such as the number of layers (from 1 to 3) and the kernel size in the convolution layers (from 3 to 5). During the training, we used the batch of size 1, with each sample consisting of 12 images.

\section{Results} Table~\ref{tab:results} summarizes the results of the experiments, with the DICE coefficient (calculated on the same testing set) being the measure of the multi-class segmentation performance.
% \vspace{-1.5cm}
% \begin{center}
\begin{table*}
\centering \caption{\textbf{Comparison with state-of-the-art methods for  multiple-region semantic segmentation in terms of mean Dice coefficient values on test data.}
			% Mean Dice coefficient values on test data.
		}
		\label{tab:results}
		% \phantom{~}\noindent
		\begin{tabular}{c@{\hskip 0.65in}c@{\hskip 0.65in}c@{\hskip 0.65in}c}
			\toprule
			Architecture&Inner Wall&Outer Wall&Tumor \\
	
			% \hline
			\midrule
			U-Net ~\cite{Ronneberger2015} & $0.911 \pm 0.017$ & $0.699 \pm 0.023$ & $0.587 \pm 0.049$\\
			% U-Net Original with BatchNorm &0.9129 $\pm$ 0.03 & 0.7117 & 0.6398\\
			U-Net Baseline~\cite{Dolz2018} & $0.897 \pm 0.022$ & $0.699 \pm 0.024$ & $0.638 \pm 0.048$\\
			U-Net Dilated~\cite{Dolz2018} & $0.921 \pm 0.013$  & $0.718 \pm 0.022$ & $0.663 \pm 0.052$\\
			U-Net Progressive Dilated~\cite{Dolz2018} & $0.914 \pm 0.016$ & $0.713 \pm 0.022$ & $0.656 \pm 0.050 $\\
			E-Net \cite{enet} & $0.913 \pm 0.015$ & $0.712 \pm 0.018$ & $0.674 \pm 0.045$ \\
			% \midrule
			% 3D: & & & \\
			\ \ \ \ 3D U-Net~\cite{iek2016_3dUnet} & $ 0.905 \pm 0.020 $ & $ 0.698 \pm 0.024 $ & $ 0.623 \pm 0.066 $\\
			%   \midrule
			\midrule
			Bi-LSTM U-Net ~\cite{bai2018recurrent}  & $0.914 \pm 0.019$ & $0.715 \pm 0.019$ & $0.642 \pm 0.051$ \\
			Temporal U-Net  & $0.917 \pm 0.016$ & $0.710 \pm 0.017$ & $0.646 \pm 0.043$\\
			Temporal U-Net Dilated & $0.910 \pm 0.017$ & $0.701\pm 0.021$ & $0.617 \pm 0.058$ \\
			\midrule
			U-Net models with hollow kernels: & & & \\
			\ \ \ \ A1 Config. 1.1 L1 & $ 0.905 \pm 0.020 $ & $ 0.698 \pm 0.024 $ & $ 0.623 \pm 0.066 $\\
			\ \ \ \ A1 Config. 1.1 L2 & $ 0.908 \pm 0.020 $ & $ 0.693 \pm 0.023 $ & $ 0.605 \pm 0.063 $\\
			\ \ \ \ A1 Config. 1.1 L3 & $ 0.894 \pm 0.028 $ & $ 0.714 \pm 0.023 $ & $ 0.652 \pm 0.052 $\\
			\ \ \ \ A2 Config. 1.1 & $ 0.918 \pm 0.016 $ & $ \textbf{0.725} \pm 0.020 $ & $ 0.662 \pm 0.052 $\\
			\ \ \ \ A2 Config. 2.1 & $ \textbf{0.923} \pm 0.013 $ & $ 0.724 \pm 0.022 $ & $ 0.673 \pm 0.052 $\\
			\ \ \ \ A2 Config. 2.2 & $ 0.922 \pm 0.014 $ & $ 0.714 \pm 0.022 $ & $ 0.642 \pm 0.054 $\\
			\ \ \ \ A2 Config. 2.3 & $ 0.917 \pm 0.018 $ & $ 0.721 \pm 0.023 $ & $ \textbf{0.675} \pm 0.052 $\\
			\ \ \ \ A2 Config. 3.1 & $ 0.914 \pm 0.018 $ & $ 0.709 \pm 0.020 $ & $ 0.635 \pm 0.050  $\\
			\midrule
		Bi-LSTM U-Net with hollow kernels & $ \textbf{0.936} \pm 0.011 $ &$ \textbf{0.736} \pm 0.018 $  & $ \textbf{0.712} \pm 0.037 $  \\
		%	\ \ \ \ A2 Config. 2.1 &  &\multirow{2}{*}{}  & \multirow{2}{*}{} \\
			\bottomrule
		\end{tabular}
	    \begin{tablenotes}
          \small
          \textbf{Note:} Model format ``A\#$_1$~Config.~\#$_2$.\#$_3$~L\#$_4$'', where \#$_1$ -- the approach number, \#$_2$ -- the number of layers employing hollow kernels, \#$_3$ -- the kernel type number (see Table 'Kernels', Supplement material), \#$_4$ -- the type of loss function (in A1 only).
        \end{tablenotes}
	\label{results}
\end{table*}
% \end{center}
The first part of the table presents the results of the experiments for several SOTA models, with the Dilated U-Net outperforming the other models in the bladder outer and inner wall segmentation and E-Net having higher accuracy in the tumor regions.
Adding temporal blocks~\footnote{See supplementary material for various configurations of temporal convolution blocks.} and Bi-LSTM to the U-Net model resulted in a significant accuracy improvement compared to the standard U-Net model, especially for the tumor class. Although, the combination of the Temporal block and the U-Net Dilated led to the decrease of the segmentation quality.

In our experiments with the hollow kernels via Approach 1, we obtained minor improvements relative to the baseline U-Net. 
% As described earlier, we used three different training loss functions (L1, L2, L3).
As a result of training using the loss function L1, Tiny U-Net learned to predict kernels that repeat the shape of the outer wall of the bladder.
The kernels generated in this manner have the same distribution of intensities, slightly differing in their values. 
In the process of training with L3, the kernels also represent some object resembling the bladder structure, but they largely differ in the intensity distribution. 
Empirically, we found approach 1 to support the concept of shape-mimicking kernels but not to strikingly outperform the benchmarks.
% Because we have not received a significant improvement when embedding hollow kernels in the first layer of the model, we did not train the complete architecture with 2 and 3 hollow kernel convolutions.
The second approach, however, outperformed the SOTA results by a good margin. We managed to achieve the best results for almost all configurations of the models with the different number of the hollow kernels, with the model A2 Config.2.1 demonstrating the best result among all the models we trained. 
The last row in the table displays the \emph{spatio-temporal} result of the experiment with the best proposed model and the Bi-LSTM block inserted between the last and the penultimate layers of the model, as shown in Fig. \ref{fig:unet_hollow} (A, C). With this model, we obtained the highest accuracy for the bladder segmentation challenge, while increasing the value of the quality metrics by 2-4\% \emph{for all classes}.

\section{Discussion}
% \clearpage %-- works but shifts typeset
\begin{figure*}
    \centering
    \includegraphics[width=\textwidth]{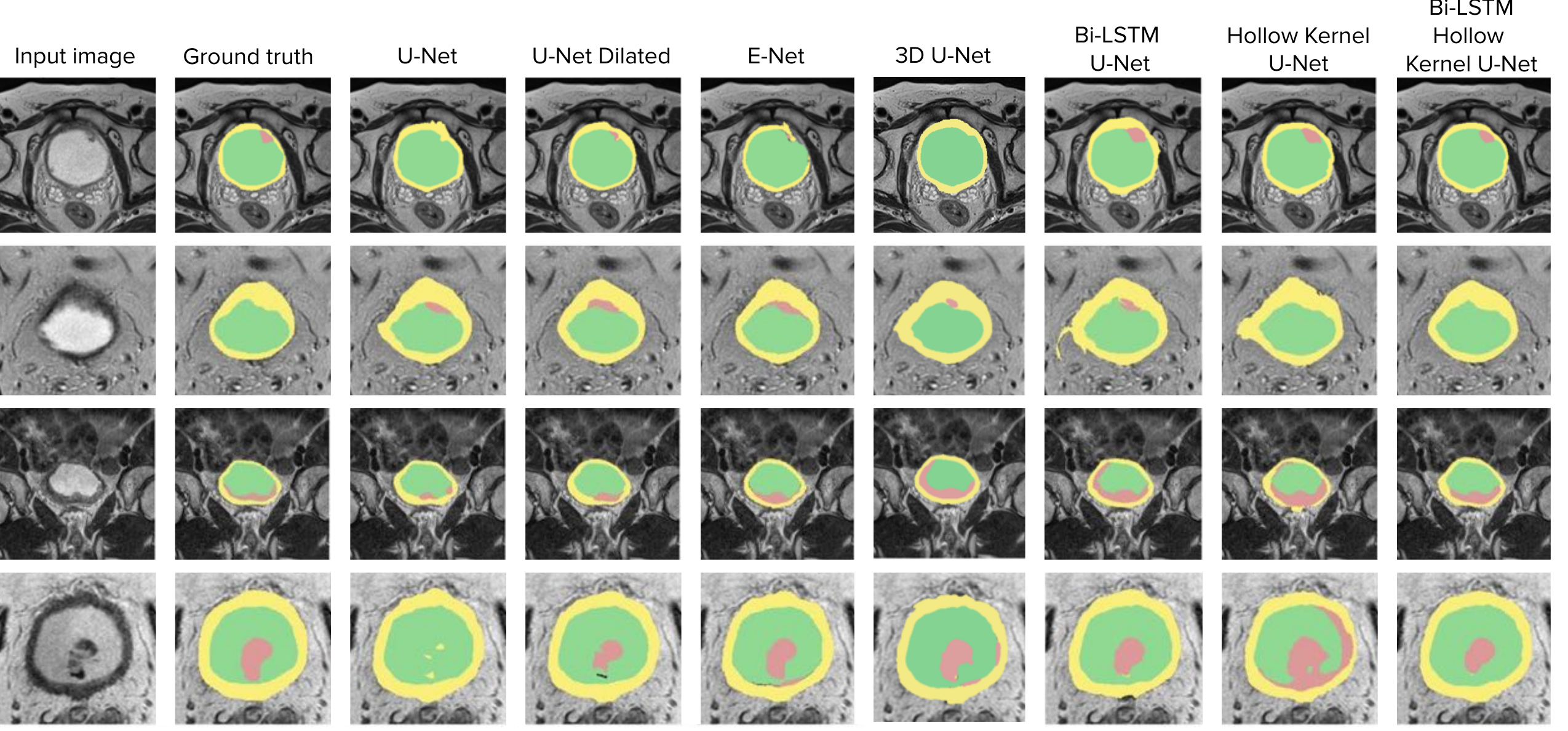}
    
    % Arxiv version
    % \includegraphics[width=\textwidth]{Figures_pdf/final_panel_v3.pdf}
    % Arxiv version  
    
    % \includegraphics[width=\textwidth]{Figures/final_panel_v3.eps}
    %\includegraphics[width=\textwidth]{Figures_small/final_panel_v3.pdf}
    \caption{\textbf{Examples of visual results achieved by key models for the most complex cases.} We follow the color encoding scheme of \cite{Dolz2018} to colorize the label map. Green and yellow regions are the inner and the outer bladder walls, respectively. Tumor regions are shown in red.} 
    \label{segm_panel}
\end{figure*}

Efficacy of the proposed models could be visually examined in Fig.~\ref{segm_panel} where we present several baseline models, the best model with the hollow kernels (U-Net with hollow kernels A2 Config. 2.1), and the same model further boosted by the Bi-LSTM block (Bi-LSTM U-Net with hollow kernels A2 Config. 2.1), which we denote as Hollow Kernels U-Net and Bi-LSTM Hollow Kernels U-Net respectively.

The first example shows a case where the bladder border is not clearly delineated and a tumor region is attached to the bladder wall. In this complex example, our models with the hollow kernels were able to accurately segment the outer wall as well as to identify the border-bound neoplastic tissues.

The second example is the case where the bladder wall is complex and has no distinct boundary. Note that all baseline models define a part of the blurred wall as a tumor area. A model with the hollow kernels handles such cases much better, however, still predicts the wrong outer contour of the wall. Finally, the proposed model with the Bi-LSTM improves the prediction of the previous model and provides accurate borders of the wall, exactly repeating morphology of the ground truth mask.

The third and the fourth examples represent clinically widespread cases of tumors with complex disintegrated structures. In one case, the tumor is attached to the inner bladder wall and partially repeats its shape; in the other case, the cancer also has a complex shape of uneven distribution of intensity. In these examples, conventional 2D and 3D models do not perform well, however, the use of Bi-LSTM helps extract the inter-slice context and provides additional information, useful for precise detection of the contours of all classes of tissues.

Our models with the hollow kernels can accurately segment bladder wall even in the cases where the bladder wall is deformed or stretched, resulting in accurate detection of tissues of complex shape and having blurred boundaries. Addition of temporal inter-slice information via the Bi-LSTM block can significantly improve the quality of tumor segmentation, regardless of its complexity, location, and shape.
\section{Conclusions} We proposed the first pipeline for constructing convolutional neural networks, where the shape of the object of interest would determine the shape of the very kernel deployed in the convolutional layers. Such learnable hollow kernels enabled us to perform precise segmentation of anatomical structures, with the bladder receiving the main focus in this report. Our model establishes a new SOTA baseline in the bladder segmentation challenge -- a particularly difficult and a previously dismissed problem in medical vision. We presented an extensive line of experiments studying the effect of these kernels on detecting multiple classes of tissues, including their scale, size, and initialization manner. The most efficient approach entails optimization of hyperparameters, with preserving the spatial structure of the hollow structure of the kernel and enhancing the spatial outcome with the temporal Bi-LSTM block. The results pave the way towards other domain-specific deep learning applications where the segmented object shape could be used to form a proper convolution kernel.

\section{Acknowledgements} We express our gratitude to the staff of the P.A. Hertzen Moscow Oncology Research Institute for providing the rare dataset.

{\small
\bibliographystyle{ieee_fullname}
\bibliography{main}
}

% --------------------------------
% SUPPLEMENTARY
% --------------------------------
\clearpage

\section{Supplementary material}
This supplementary material is structured as follows.
In Section~\ref{s:dataset}, we present descriptive statistics of the dataset acquired at the P. Hertsen Moscow Oncology Research Institute~\footnote{The annotated dataset will be made public following the ongoing regulatory approval. We stress that there are no other bladder datasets publicly available today. Bladder segmentation is one of the most dismissed problems in the community.}.
In Section~\ref{s:hkce}, we present examples of the simple hollow kernel structures that perform optimal segmentation of hollow objects.
We then present graphs that showcase the performance of our models in comparison with the other state-of-the-art baselines on different datasets. Next, we present some examples to show how deftly our models handle the challenges posed by partial object occlusions, intensity variations, and the viewpoint alterations. 
We proceed by showcasing temporal convolutions approach performance in Sections~\ref{s:stm} and ~\ref{s:res_add}.
\section{Data}\label{s:dataset}
% DATA-TABLE
The patient data employed in the current study are presented in Table \ref{tab:cdata}, featuring clinical statistics. These data were annotated by two expert in urologic oncology.
\begin{table}[h]
    \caption{Clinical characteristics of bladder cancer patients employed in the study.}
    \centering
    \begin{tabular}{cc}
        \toprule
        Characteristic & Value \\
        \midrule
        Patients, \# (\%) & 32 (100\%) \\
        Female, \# (\%)  & 3 (9.68\%) \\
        Male, \# (\%)  & 29 (90.32\%) \\
        Age, Median (Range), yrs & 62 [38, 88]\\
        \bottomrule
    \end{tabular}
    \label{tab:cdata}
\end{table}

%-----------------------
\section{Intuition behind hollow kernels on synthetic data}\label{s:hkce}

In this section, we present a variety of examples of how hollow kernel convolution works on hollow synthetic objects in addition to Section 3 of the main text.

\begin{figure*}
    \centering
    \includegraphics[width=\textwidth]{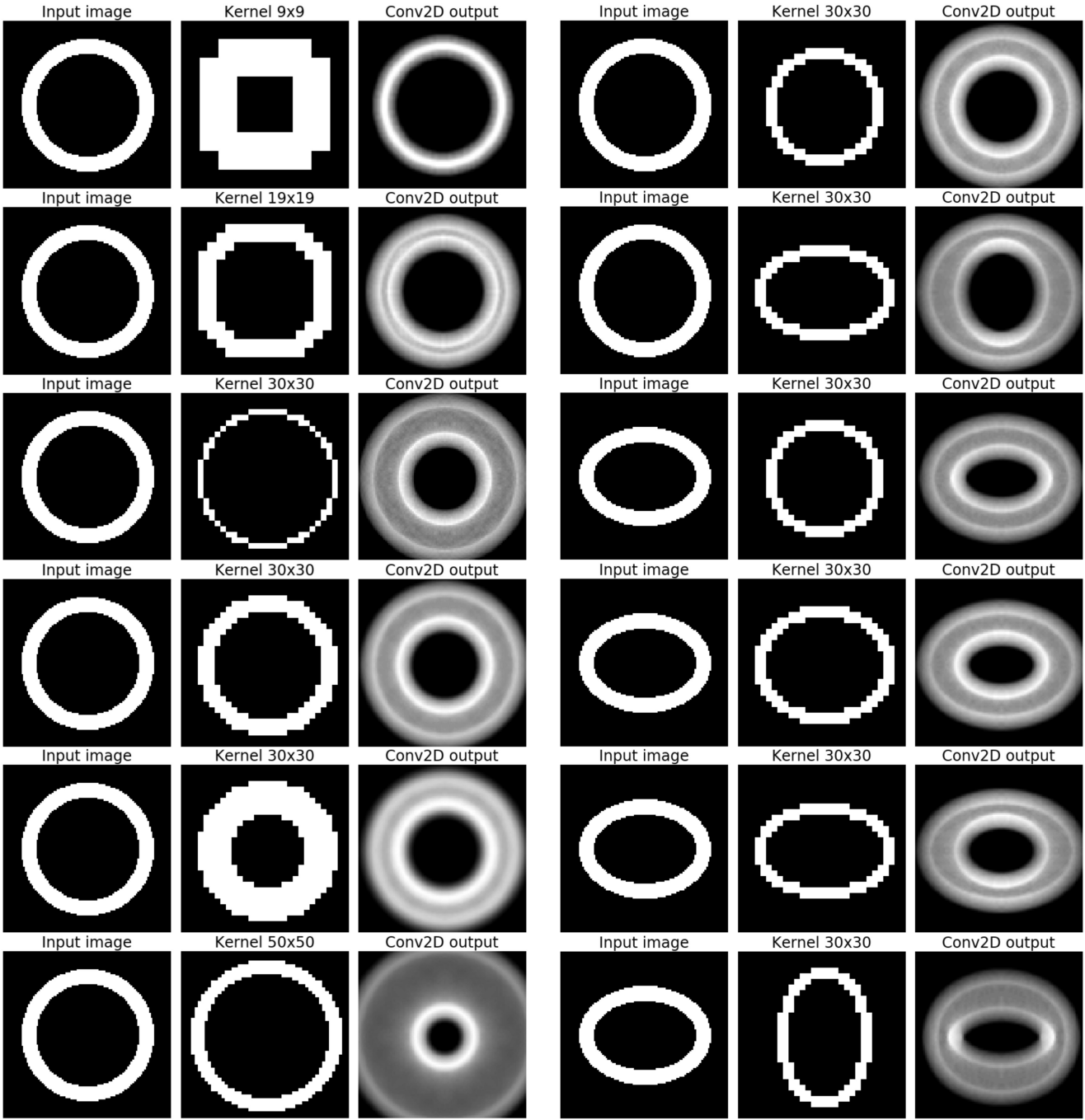}
    \caption{Examples of the use of hollow kernels of different shapes and scales in convolutions on hollow synthetic objects. Notice how images convolved with a same-shape object emphasize the borders, being the motivation behind this study.}
    \label{app:conv_aug_big}
\end{figure*}

Let us consider the hollow object of a simple form. In the left column of Fig.~\ref{app:conv_aug_big}, there are several generated hollow objects in a 100 $\times$ 100 pixel image. Applying convolution to such an object with the kernel also a hollow structure, one will see an interesting effect -- the borders of the original object will be highlighted. This effect is most noticeable on the middle lines of the image, where the kernel and the object in the input image have the same bagel shape.

We observe the same effect for the ellipse shape (see examples in the left side of the Fig.~\ref{app:conv_aug_big}). The boundaries of the resulting object are also sharply outlined.  However, the resulting object is deformed relative to the original one. The walls of the object are thickened in the direction coinciding with the direction of the highest curvature of the kernel.

Note that the ratio between the linear kernel size and the width of the input hollow object wall plays a key role in this effect. If the linear kernel size is much smaller than the wall thickness, the usual blurring of the image is observed (see the top left line in Fig.~\ref{app:conv_aug_big}).
When the kernel size is increased, a thin hollow structure appears in the output image. When the spatial size of the kernel becomes comparable to the width of the wall, the boundary of the resulting object coincides with the boundary of the original one. With the further expansion of the kernel, the width of the output element wall is also growing.
\begin{figure}
    \centering
         \centering
         \includegraphics[width=0.40\textwidth]{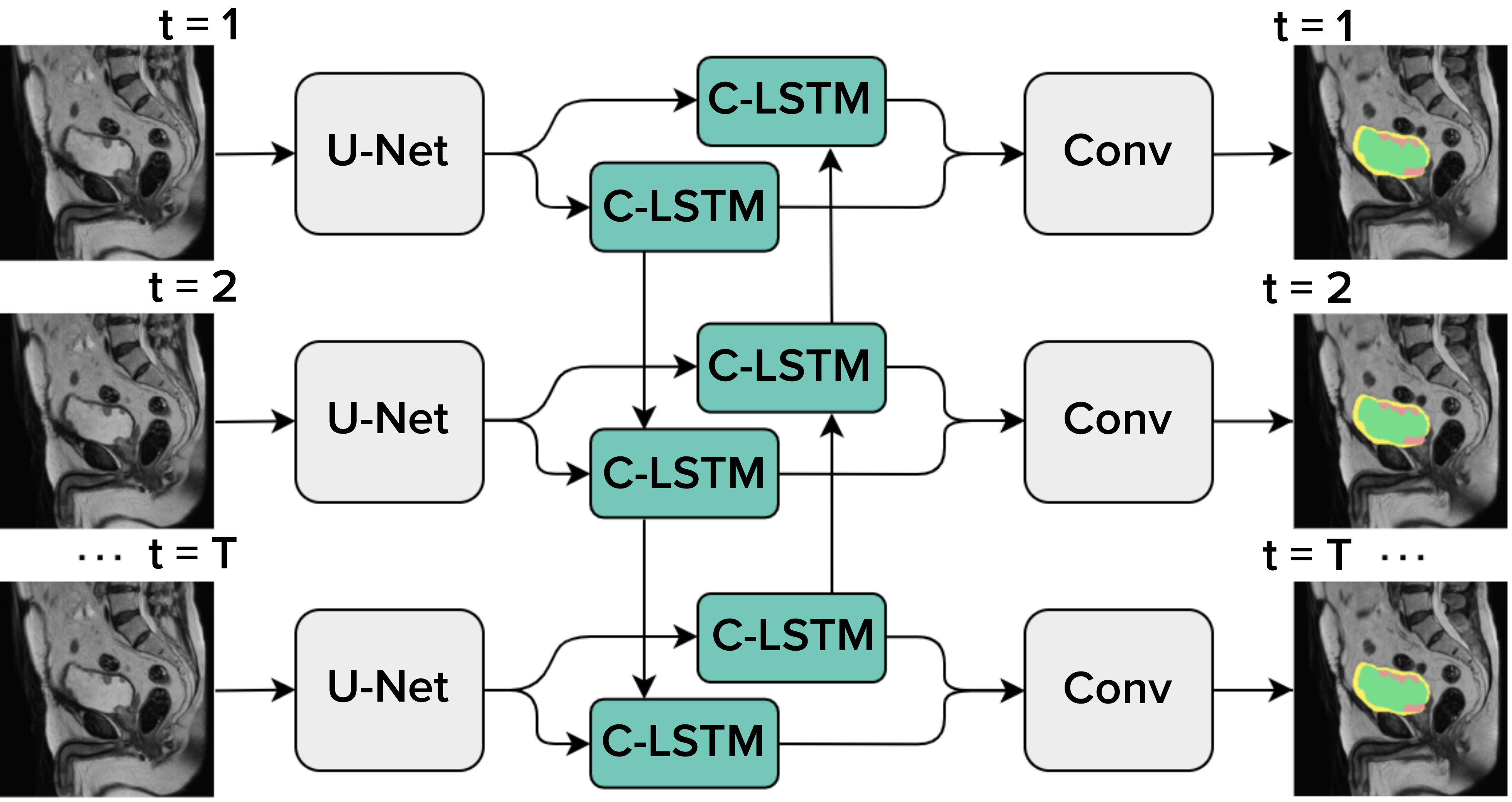}
         \caption{Bi-LSTM U-Net model}
        %  \label{fig:model_bilstm}
    \hspace{0.5em}

    % \caption{\textbf{The scheme of proposed \textit{spatio-temporal} models.} The input sequence is fed through U-Net-based model (U-Net Dilated or Hollow Kernels U-Net). The feature maps, extracted from the second last layer of U-Net, are inherited into the block responsible for retrieving the global context.
    % Finally, it predicts the label map sequence.}
    \label{fig:model_bilstm}
\end{figure}

\begin{figure}
    \centering
    \includegraphics[width=0.40\textwidth]{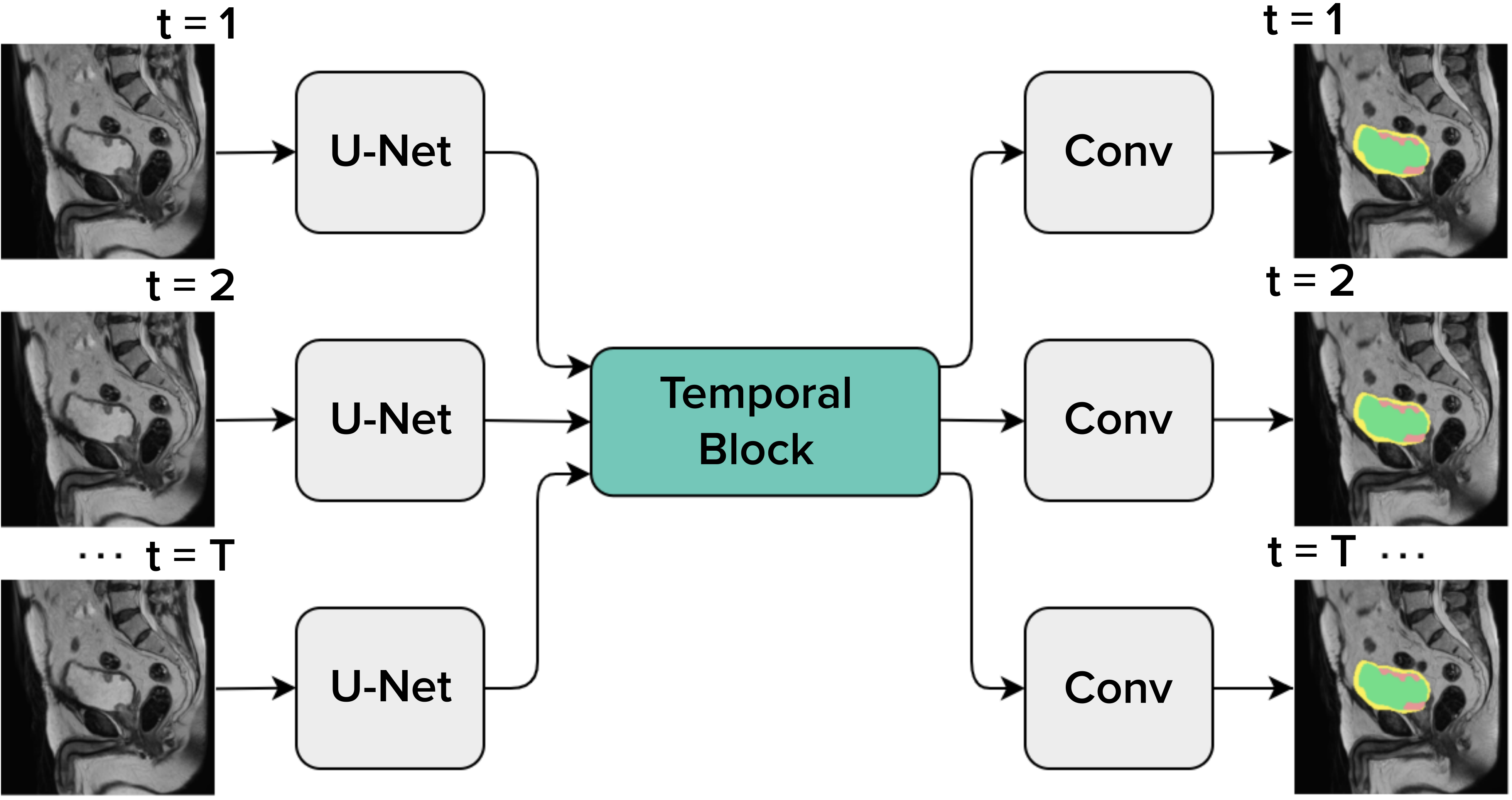}
    \caption{Temporal U-Net model}
    \label{fig:model_temporal}
\end{figure}
%-----------------------
\section{Spatial models: Hollow Kernel U-Net Architectures}\label{hkua}

As described in the Section 4 of the main text, we used different U-Net and U-Net Dilated configurations. The parameters of the first three convolutions of such models are presented in Table \ref{tab:three_blocks}. We refer the reader to this table to explain the notation ``A\#$_1$~Config.~\#$_2$.\#$_3$~L\#$_4$'' from the main text.

\begin{table*}[!ht]
    \centering
    % \begin{threeparttable}

        \caption[Parameters of the first three convolution blocks of alternative models explored.]{Parameters of the first three convolution blocks of alternative models explored.} \label{tab:three_blocks}

        \begin{tabular}{cc@{\hskip 0.35in}c@{\hskip 0.35in}c@{\hskip 0.35in}c@{\hskip 0.35in}c@{\hskip 0.35in}c@{\hskip 0.35in}c@{\hskip 0.35in}c@{\hskip 0.35in}c@{\hskip 0.35in}c@{\hskip 0.35in}c@{\hskip 0.35in}}
    
            & & \mcrot{1}{c@{\hskip 0.48in}}{90}{U-Net} & \mcrot{1}{c@{\hskip 0.48in}}{90}{A1 Config. 1.1} & \mcrot{1}{c@{\hskip 0.48in}}{90}{U-Net Dilated} & \mcrot{1}{c@{\hskip 0.48in}}{90}{A2 Config. 1.1} & \mcrot{1}{c@{\hskip 0.48in}}{90}{A2 Config. 2.1} & \mcrot{1}{c@{\hskip 0.48in}}{90}{A2 Config. 2.2} &  
            \mcrot{1}{c@{\hskip 0.48in}}{90}{A2 Config. 2.3} & 
            \mcrot{1}{c@{\hskip 0.48in}}{90}{A2 Config. 3.1} & \mcrot{1}{c@{\hskip 0.48in}}{90}{A2 Config. 3.2} &  
            \mcrot{1}{c@{\hskip 0.48in}}{90}{A2 Config. 3.3} \\
        % \midrule \midrule
            \\
            % \hline
            \hline
             \parbox[t]{2mm}{\multirow{5}{*}{\rotatebox[origin=c]{90}{Conv 1.1}}} 
             & channels in  & 1 & 1 & 1 & 1 & 1 & 1 & 1 & 1 & 1 & 1  \\
             & channels out & 32 & 32 & 32 & 32 & 32 & 32 & 32 & 32 & 32 & 32\\
             &  kernel  & 3 & \textbf{20} & 3 & \textbf{10} & \textbf{10} & 
             \textbf{10} & \textbf{10} & \textbf{10} & \textbf{10} & \textbf{10}\\
             & dilation & 1 & 1 & 1 & 1 & 1 & 1 & 1 & 1 & 1 & 1\\
             & stride & 1 & 1 & 1 & 1 & 1 & 1 & 1 & 1 & 1 & 1\\
             \hline
             \parbox[t]{2mm}{\multirow{5}{*}{\rotatebox[origin=c]{90}{Conv 1.2}}} 
             & channels in & 32 & 32 & 32 & 32 & 32 & 32 & 32 & 32 & 32 & 32\\
             & channels out & 32 & 32 & 32 & 32 & 32 & 32 & 32 & 32 & 32 & 32\\
             &  kernel  & 3 & 4 & 3 & 4 & 4 & 4 & 4 & 4 & 4 & 4\\
             & dilation & 1 & 1 & 1 & 3 & 3 & 3 & 3 & 3 & 3 & 3\\
             & stride & 1 & 1 & 1 & 1 & 1 & 1 & 1& 1 & 1 & 1\\
             \hline
             \parbox[t]{2mm}{\multirow{5}{*}{\rotatebox[origin=c]{90}{Conv 2.1}}} 
             & channels in & 32 & 32 & 32 & 32 & 32 & 32 & 32 & 32 & 32 & 32\\
             & channels out & 128 & 128 & 128 & 128 & 128 & 128 & 128 & 128 & 128 & 128\\
             &  kernel  & 3 & 3 & 3 & 3 & \textbf{10} & \textbf{20} & 
             \textbf{10} & \textbf{10} & \textbf{20} & \textbf{10}\\
             & dilation & 1 & 1 & 2 & 2 & \textbf{2} & \textbf{1} & \textbf{1} & \textbf{2} & \textbf{1} & \textbf{1}\\
             & stride   & 1 & 1 & 2 & 2 & 2 & 2 & 2 & 2 & 2 & 2\\
             \hline
             \parbox[t]{2mm}{\multirow{5}{*}{\rotatebox[origin=c]{90}{Conv 2.2}}} 
             & channels in & 128 & 128 & 128 & 128 & 128 & 128 & 128 & 128 & 128 & 128\\
             & channels out & 128 & 128 & 128 & 128 & 128 & 128 & 128 & 128 & 128 & 128\\
             &  kernel  & 3 & 3 & 3 & 3 & 4 & 4 & 4 & 4 & 4 & 4\\
             & dilation & 1 & 1 & 2 & 2 & 3 & 3 & 3 & 3 & 3 & 3\\
             & stride   & 1 & 1 & 1 & 1 & 1 & 1 & 1 & 1 & 1 & 1\\
             \hline
             \parbox[t]{2mm}{\multirow{5}{*}{\rotatebox[origin=c]{90}{Conv 3.1}}} 
             & channels in & 128 & 128 & 128 & 128 & 128 & 128 & 128 & 128 & 128 & 128\\
             & channels out & 256 & 256 & 256 & 256 & 256 & 256 & 256 & 256 & 256 & 256\\
             &  kernel & 3 & 3 & 3 & 3 & 3 & 3 & 3 & \textbf{10} & \textbf{40} & \textbf{10}\\
             & dilation & 1 & 1 & 4 & 4 & 4 & 4 & 4 & \textbf{4} & \textbf{1} & \textbf{1}\\
             & stride & 1 & 1 & 2 & 2 & 2 & 2 & 2 & 2 & 2 & 2\\
             \hline
             \parbox[t]{2mm}{\multirow{5}{*}{\rotatebox[origin=c]{90}{Conv 3.2}}} 
             & channels in & 256 & 256 & 256 & 256 & 256 & 256 & 256 & 256 & 256 & 256\\
             & channels out & 256 & 256 & 256 & 256 & 256 & 256 & 256 & 256 & 256 & 256\\
             &  kernel & 3 & 3 & 3 & 3 & 3 & 3 & 3 & 3 & 3 & 3\\
             & dilation & 1 & 1 & 4 & 4 & 4 & 4 & 4 & 4 & 4 & 4\\
             & stride & 1 & 1 & 1 & 1 & 1 & 1 & 1 & 1 & 1 & 1\\
             \hline
        \end{tabular}
		\begin{tablenotes}
			\small
			\textbf{Note:} Model format ``A\#$_1$~Config.~\#$_2$.\#$_3$~L\#$_4$'', where \#$_1$ -- the approach number, \#$_2$ -- the number of layers employing hollow kernels, \#$_3$ -- the kernel type number (see Table 'Kernels', Supplement material), \#$_4$ -- the type of loss function (in A1 only).
		\end{tablenotes}        
    %     \begin{tablenotes}
    %       \small
    %       \textbf{Note:} Model format ``A\#$_1$~Config.~\#$_2$.\#$_3$~L\#$_4$'', where \#$_1$ -- the approach number, \#$_2$ -- the number of layers employing hollow kernels, \#$_3$ -- the kernel type number (see Table 'Kernels', Supplement material), \#$_4$ -- the type of loss function (in A1 only).
    %     \end{tablenotes}
    % \end{threeparttable}
\end{table*}

%-----------------------
\section{Spatio-Temporal models}\label{s:stm}

The final model, described in Sections 3 and 4 of the main text, is the model that combines the Hollow kernel U-Net and Bi-Directional Convolutional LSTM block (see Fig. \ref{fig:model_bilstm}, where U-Net blocks represent any type of U-Net model).

As baselines we used the same recurrent architecture with the original U-Net Dilated as well as the model with a the Temporal convolutional block inserted into it. In more details, the Temporal U-Net  (Fig.~\ref{fig:model_temporal}) retrieves the second last layer of U-Net as feature maps of input image sequence, and then connect them using temporal convolutions and finally predicts the probability maps. Feature maps extracted on the penultimate layer contain low-level information about the pixels of the input image. The exchange of information between such feature maps across the temporal domain ensures the refinement of these feature maps.
% \label{s:stm}

\begin{figure*}
    \centering
    \includegraphics[width=\textwidth]{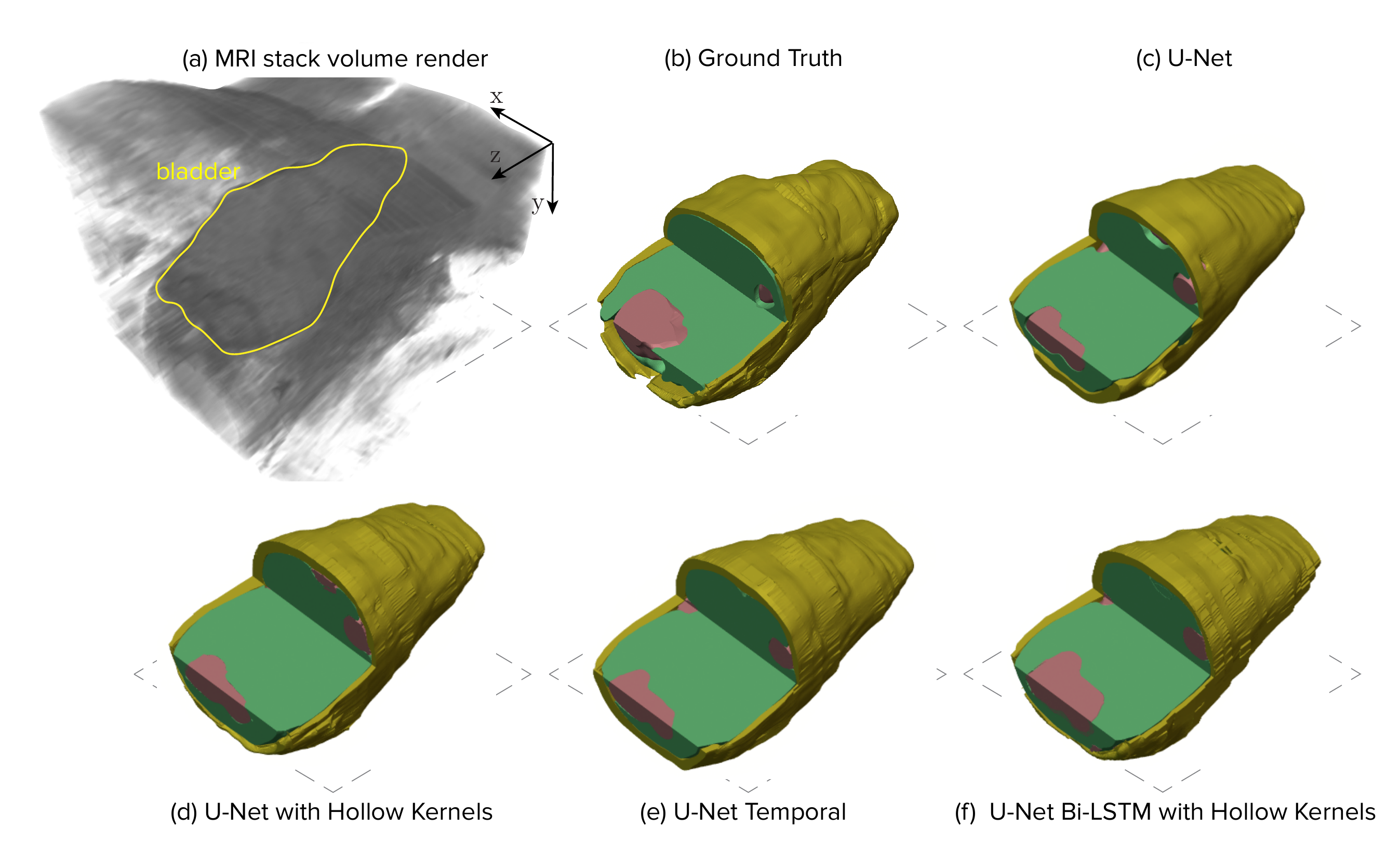}
    \caption{Results of segmentation in 3D. Notice how the \emph{spatio-temporal} case extracts all anatomical structures with best accuracy for each class of tissue. Quantitative metrics are given in the main text.} 
    \label{fig:promo_multi}
\end{figure*}
% \newpage
\section{Results}\label{s:res_add}
\begin{figure*}
    \centering
    \includegraphics[width=0.89\textwidth]{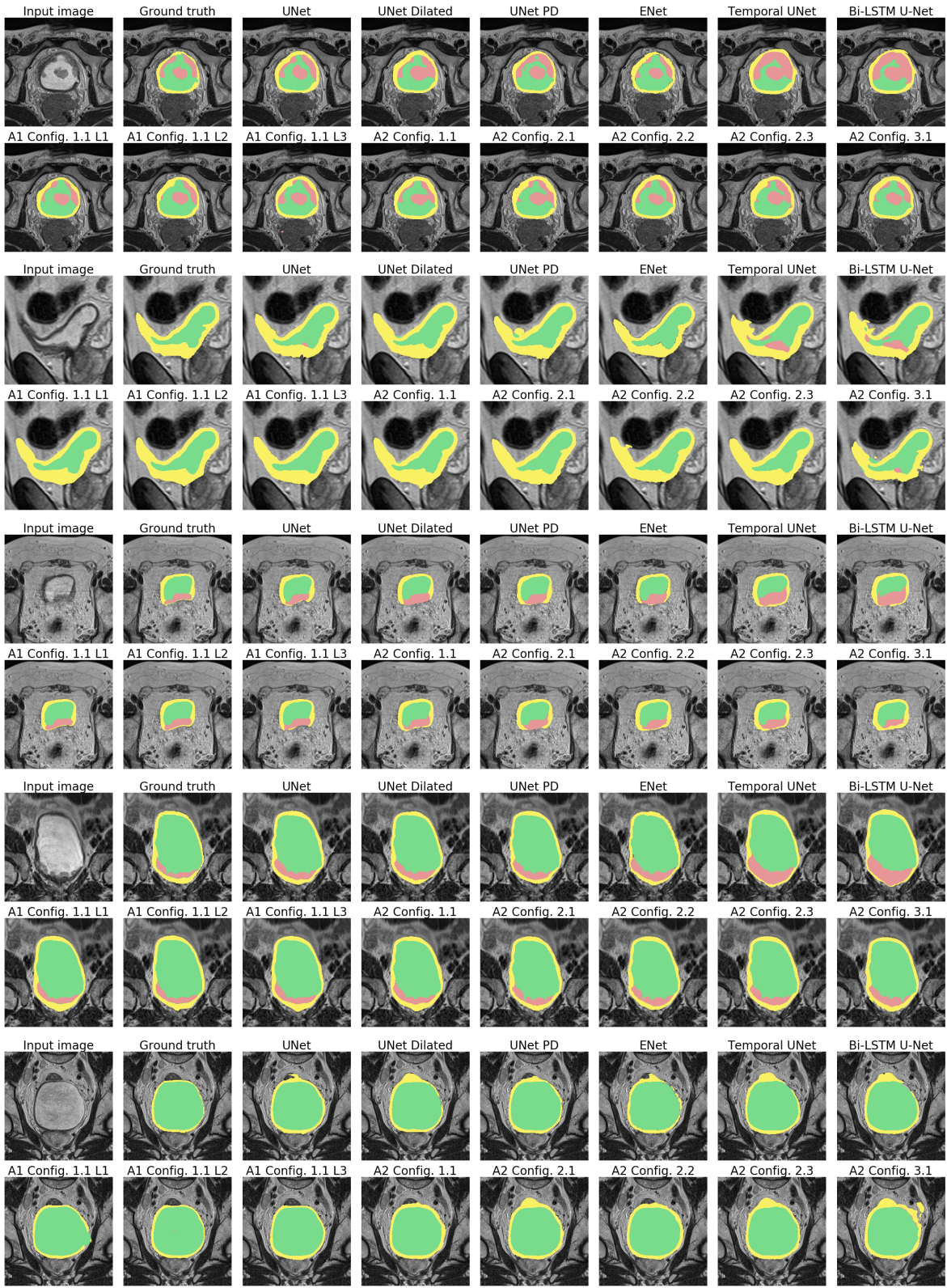}
    \caption{Examples of performance of all models. Green and yellow regions are the inner and the outer bladder walls, respectively. Tumor regions are shown in red.} 
    \label{fig:segm_examp}
\end{figure*}

\begin{figure*}
    \centering
    \begin{subfigure}{\textwidth}
         \centering
         \includegraphics[width=\textwidth]{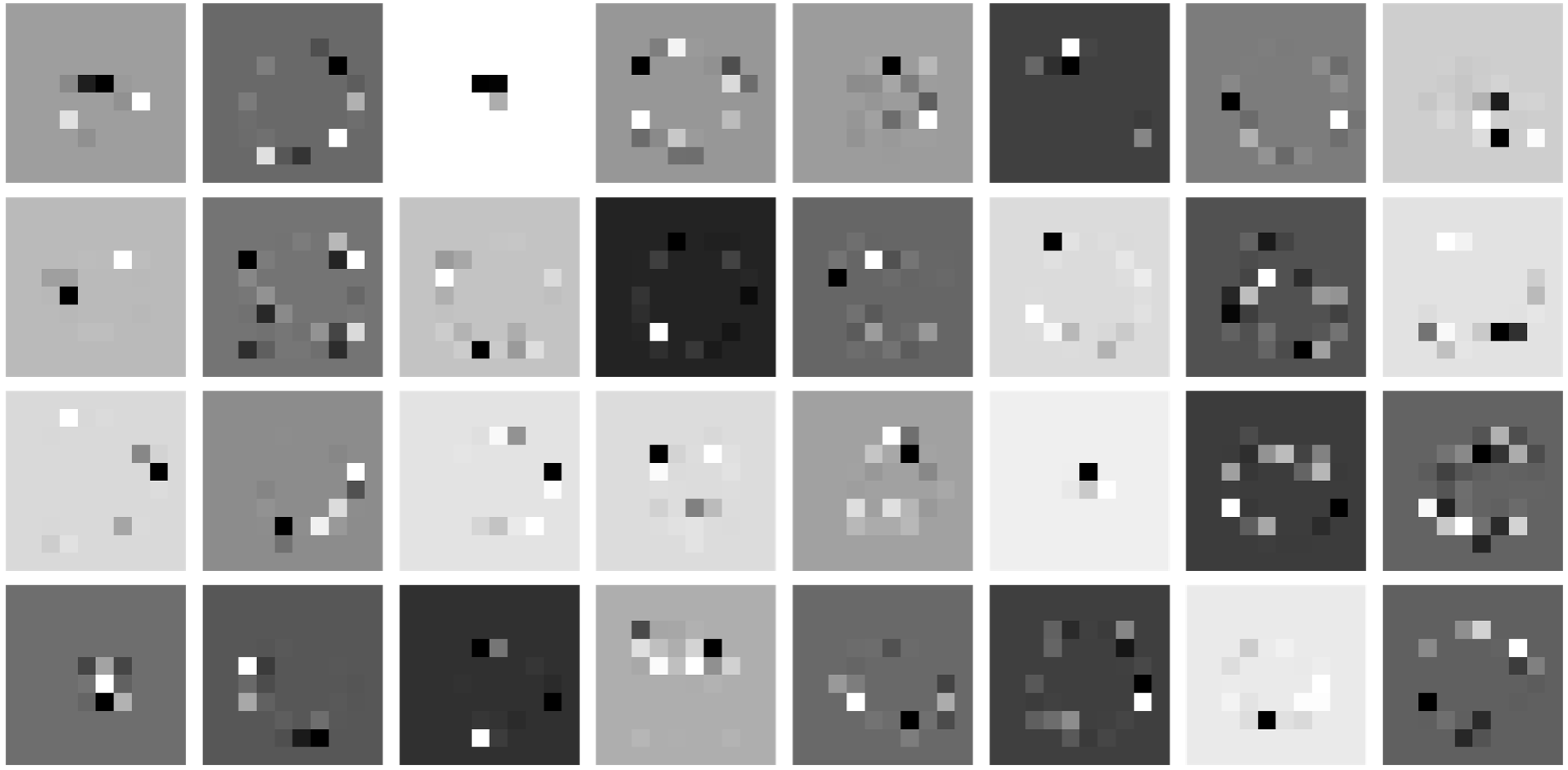}
         \caption{Initialization of hollow kernels for the first convolution layer.}
    \end{subfigure}

    \begin{subfigure}{\textwidth}
         \centering
         \includegraphics[width=\textwidth]{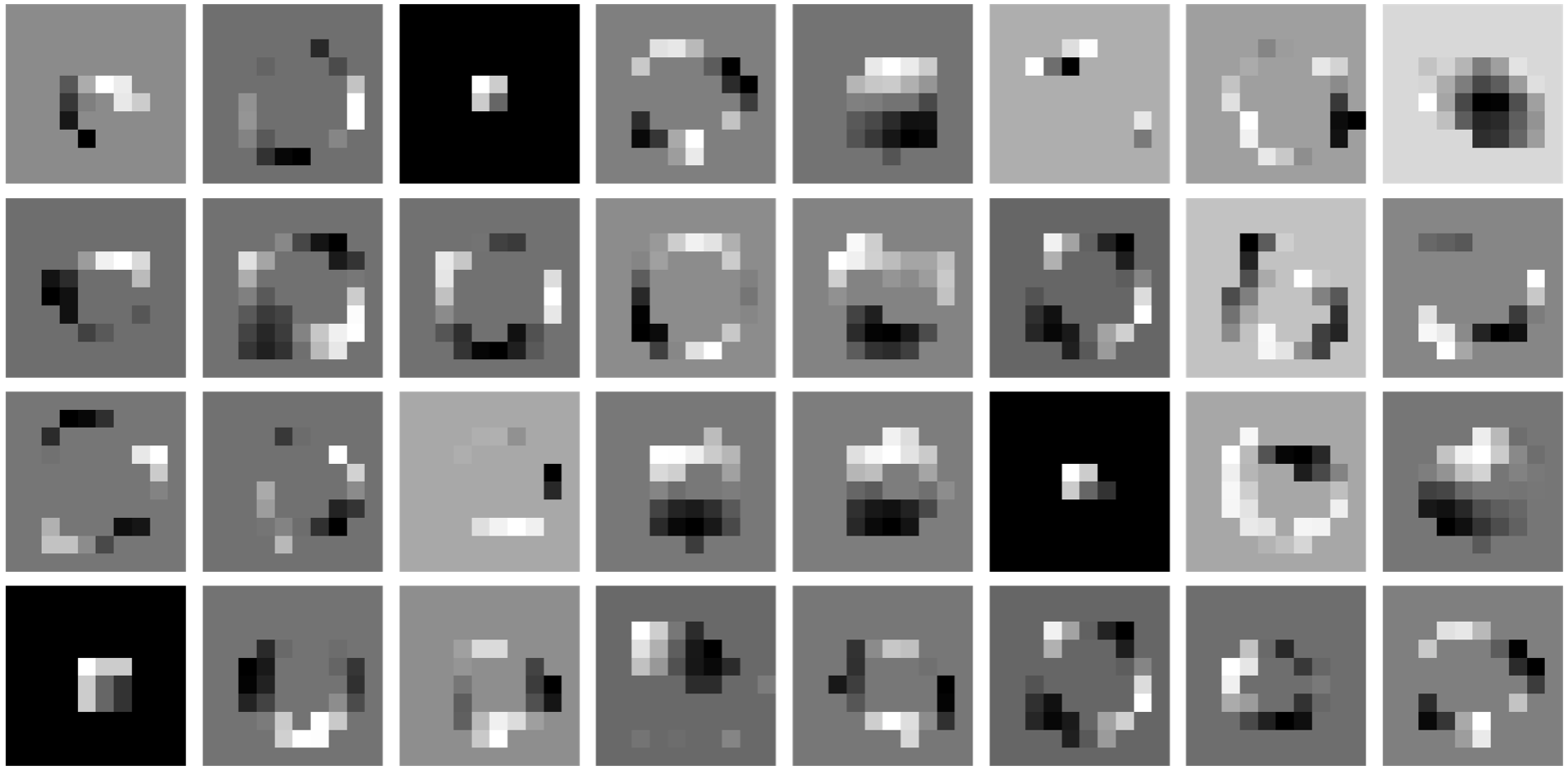}
         \caption{Trained kernels of the first convolution layer.}
    \end{subfigure}
    \caption{Kernels of the first convolution layer of the proposed model in approach 2 (\textbf{A2}), where the training process is performed preserving the structure of the kernels. Notice the learnt hollow bladder-like contours of the kernels.}
    \label{fig:kernels_init}
\end{figure*}

In this section, we present details on the results presented in the main paper, focusing on the visualization of each model considered. 
Fig. \ref{fig:segm_examp} presents examples of segmentation performed by different models. For each example, the first line contains the input image, the ground truth and predictions of baseline models. The second line represents the predictions of the best configurations of U-Net model with hollow kernels, described in Section \ref{hkua}. 
Then, Fig. \ref{app:examples_a1_config_11_L1} and \ref{app:examples_a1_config_11_L3} show a few examples of predictions obtained with Hollow Kernels U-Net models and employed kernels generated by the Tiny U-Net.
Fig. \ref{fig:kernels_init} represents the convolution kernels used before and after training while preserving the hollow kernel structure.

\begin{figure*}

     \centering
     \includegraphics[width=\textwidth]{Figures_pdf/app_example_a1_config_11_L1.pdf}

     \caption[A few examples of the performance of the U-Net model with one convolution layer with kernels trained to resemble the shape of bladder (A1 Config. 1.1 L1).]{Examples of the performance of the U-Net model with one convolution layer with kernels trained to resemble the shape of bladder (A1 Config. 1.1 L1). \\\textit{Left column:} the input image, ground truth, and main model prediction are presented. Green and yellow regions are the inner and outer bladder walls, respectively. Tumor regions are shown in red. \\\textit{Right image mosaic:} these are the 32 kernels of size 20 $\times$ 20 generated for this image by Tiny-U-Net and used within the convolution layer of the main model.}
     \label{app:examples_a1_config_11_L1}
\end{figure*}

% \newpage
\begin{figure*}
     \centering

     \includegraphics[width=\textwidth]{Figures_pdf/app_example_a1_config_11_L3.pdf}
    
     \caption{Examples of the performance of the U-Net model with a single convolution layer, where the kernels are trained so that the sum of them gives the shape of the bladder (A1 Config. 1.1 L3). \\\textit{Left column:} the input image, ground truth, and the main model prediction. Green and yellow regions are the inner and outer bladder walls, respectively. Tumor regions are shown in red. \\\textit{Right image mosaic:} these are the 32 kernels of size 20 $\times$ 20 generated for this image by Tiny-U-Net and used within the convolution layer of the main model.}
     \label{app:examples_a1_config_11_L3}
\end{figure*}

% \newpage

% --------------------------------

\end{document}